\documentclass[10pt,twocolumn,letterpaper]{article}

\usepackage{iccv}
\usepackage{times}
\usepackage{epsfig}
\usepackage{graphicx}
\usepackage{amsmath}
\usepackage{amssymb}

\usepackage[utf8]{inputenc}
\usepackage[T1]{fontenc}
\usepackage{textcomp}
\usepackage{gensymb}

\usepackage{enumerate}

\usepackage{multirow}
\usepackage{tabu}
\usepackage{subfigure}

\DeclareMathAlphabet{\mathpzc}{OT1}{pzc}{m}{it} %

\DeclareMathOperator*{\argmin}{argmin}

\newcommand{\trans}[1]{{#1}^{\ensuremath{\mathsf{T}}}} 

\usepackage[pagebackref=true,breaklinks=true,letterpaper=true,colorlinks,bookmarks=false]{hyperref}

\iccvfinalcopy 

\def\iccvPaperID{888} 

\ificcvfinal\fi
\begin{document}

\title{Towards Arbitrary-View Face Alignment by Recommendation Trees\thanks{This is our original submission to ICCV 2015.}}

\author{Shizhan Zhu$^1$ \quad Cheng Li$^2$  \quad Chen Change Loy$^{1,3}$  \quad Xiaoou Tang$^{1,3}$\\
$^1$Department of Information Engineering, The Chinese University of Hong Kong \\ $^2$SenseTime Group\\ $^3$Shenzhen Institutes of Advanced Technology, Chinese Academy of Sciences \\
{\tt\small zs014@ie.cuhk.edu.hk, chengli@sensetime.com, ccloy@ie.cuhk.edu.hk, xtang@ie.cuhk.edu.hk}
}

\maketitle

\begin{abstract}
Learning to simultaneously handle face alignment of arbitrary views, \eg~frontal and profile views, appears to be more challenging than we thought.
The difficulties lay in i) accommodating the complex appearance-shape relations exhibited in different views, and ii) encompassing the varying landmark point sets due to self-occlusion and different landmark protocols.
Most existing studies approach this problem via training multiple viewpoint-specific models, and conduct head pose estimation for model selection. This solution is intuitive but the performance is highly susceptible to inaccurate head pose estimation.
In this study, we address this shortcoming through learning an Ensemble of Model Recommendation Trees (EMRT), which is capable of selecting optimal model configuration without prior head pose estimation.
The unified framework seamlessly handles different viewpoints and landmark protocols, and it is trained by optimising directly on landmark locations, thus yielding superior results on arbitrary-view face alignment.
This is the first study that performs face alignment on the full AFLW dataset with faces of different views including profile view. State-of-the-art performances are also reported on MultiPIE and AFW datasets containing both frontal- and profile-view faces.

\end{abstract}

\section{Introduction}
\label{intro}
\begin{figure}
  \centering
  \includegraphics[width=1.00\linewidth]{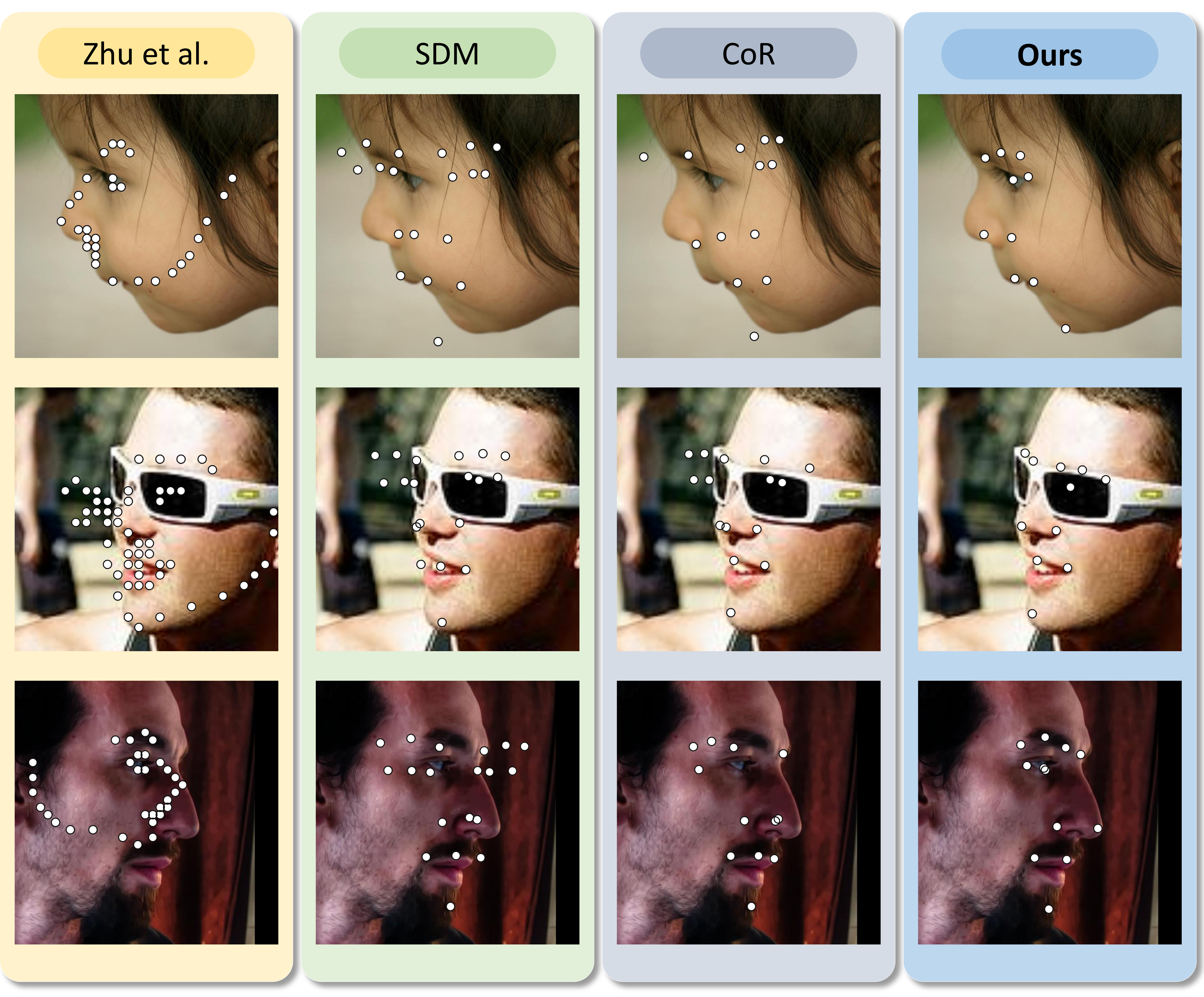}\\
  \caption{Results on four methods (from left to right): \cite{zhu2012face} models facial structure by mixture of trees; SDM \cite{xiong2013supervised} is a representative discriminative regression method that achieves state-of-the-art performance on frontal-face-biased datasets~\cite{belhumeur2011localizing, le2012interactive}; CoR \cite{yu2014consensus} is specially proposed to handle external-object occlusion; and our proposed method based on model recommendation.}
  \label{fig:intro}
\end{figure}

Face alignment is essential for many facial analysis tasks such as facial expression detection~\cite{du2014compound, zen2014unsupervised}, attribute recognition~\cite{datta2011hierarchical}, face image processing~\cite{yang2013structured}, face recognition and verification~\cite{taigman2014deepface, chen2013automatic, sun2014deep, dou2014benchmarking, hermosilla2011thermal}.
Significant improvements have been achieved in face alignment with the emergence of discriminative regression methods~\cite{xiong2013supervised, ren2014face, kazemi2014one}.
Nevertheless, most of these methods tend to confine their solutions on faces with limited view range, in which at least both two eyes are visible and all landmarks can be explicitly annotated. Some studies~\cite{zhou2005bayesian,xing2014towards} can handle multi-view face alignment, but they only reported results on faces with $\sim\pm$45$\degree$ viewpoint variation.
When view changes dramatically, these models often produce erroneous shape estimation (Fig. \ref{fig:intro}).
%
%
%
However, non-frontal face constitutes a considerable fraction of faces found in various sources, such as web images, social network photos, and surveillance videos.
Considering alignment only for near-frontal view severely limits the scalability of many subsequent facial analyses in real-world environment.
In this paper, we wish to study the limitations of using a single discriminative regression model~\cite{xiong2013supervised, ren2014face, kazemi2014one}, and formulate a unified model recommendation framework to handle arbitrary view poses up to profile view $\sim\pm$90$\degree$.

Simultaneously considering face alignment for arbitrary view poses is non-trivial:
\textit{i)} For large view variation some landmarks such as those on half of the face might become invisible due to self-occlusion. Solving this problem requires not only the capability to infer the visibility of occluded points, but also the ability to handle varying landmark point sets. For instance, MultiPIE has 39 points for profile-view and 68 points for frontal-view.
\textit{ii)} Importantly, due to the inherently multimodal and complex appearance-shape relations exhibited in multi-view faces, a single model would easily fail to fit the landmarks (Fig.~\ref{fig:intro}).
The aforementioned challenges demand multiple face alignment models specially trained on different view poses, so as to handle different landmark protocols and multimodal appearance-shape relations. How to jointly and seamlessly consider all models for aligning an arbitrary face pose remains an understudied problem.

There are a few attempts~\cite{dantone2012real,zhao2014unified} to address the arbitrary-view face alignment.
These methods first estimate the head pose and subsequently choose the corresponding alignment model from a model pool for landmark estimation.
Nonetheless, head pose estimation in unconstrained settings itself is a challenging task. Existing methods mostly rely on global features for estimation, which can be susceptible to illumination, occlusion, and background clutter. Error in head pose estimation inevitably affects the performance of subsequent face alignment.
Approaches have also been proposed to handle occlusion caused by external objects such as hands, hairs, or sunglasses~\cite{burgos2013robust,yu2014consensus}. Nonetheless, there are intrinsic differences between the handling of external occlusion and self-occlusion caused by view variation.
In particular, given a specific pose, the underlying shape constraint remains unchanged in the presence of external occlusion. Even if some facial parts are occluded, accurate landmarks can still be inferred using a single model. On the contrary, self-occlusion caused by view changes is a more challenging problem.
Our experiments show that a direct application of face alignment specifically designed for handling external occlusion fails to cope with profile-view poses (CoR~\cite{yu2014consensus} in Fig. \ref{fig:intro}).

Handling face alignment with different view poses requires a framework that can seamlessly handle different face models and landmark protocols from frontal view to profile view, and select optimal model configuration given an arbitrary pose.
To this end, we cast the problem as a model recommendation problem and propose a framework called Ensemble of Model Recommendation Trees (EMRT).
Decision forests \cite{breiman2001random} have been extensively used in many applications including facial alignment due to its effectiveness and efficiency. In this study, we reformulate it for the purpose of model recommendation.
%
%
In the proposed framework, we first prepare a model pool by training a set of alignment models, each of which concentrates on processing specific cluster of face samples with similar face angles and landmark protocols\footnote{Note that our framework can naturally handle models encompassing different in-plane and out-of-plane rotations. In addition, the framework flexibly handles varying number of models depending on the complexity of the application scenarios, and it can be used along with different existing discriminative regression methods~\cite{xiong2013supervised, ren2014face, kazemi2014one}.}.
Given a face image with unknown pose, the EMRT takes features based on the shape responses from different models as input, and produces a model rating vector that encodes the preference on different models. Face alignment is then accomplished by aggregating the consensus of different models based on the rating.

The main contributions of this study are as follows:

\noindent
(1) We propose a novel approach that is capable of handling arbitrary pose face alignment, up to profile view. To our knowledge, this is the first study that is capable of training and evaluating on the challenging \textbf{full} Annotated Facial Landmarks in the Wild (AFLW) dataset \cite{kostinger2011annotated}~\footnote{The full AFLW dataset contains 25,993 images. Existing studies~\cite{cootes2012robust, shen2013detecting, yang2013sieving, Zhang2014} only select $\leq$25\% images with near-to-frontal views for evaluations.}. State-of-the-art performances are also achieved on MultiPIE~\cite{sim2003cmu} and AFW~\cite{zhu2012face}, which contain challenging profile-view faces.

\noindent
(2) We formulate a new forest-based recommendation framework with appealing properties.
Firstly, it has a new splitting function that jointly optimises the partitions of features and the learning of model rating vector. The splitting function is formulated such that it optimises directly the landmark locations, thus leading to accurate alignment given arbitrary poses.
Secondly, the objective function is optimised to produce a model rating vector that captures the unique correlation between models so that semantically nearer models (\eg~those trained with faces in nearby face angle range) share closer values in the model rating vector, than those further away in the model space.
Thanks to these properties, the proposed model recommendation approach is more robust than other model selection alternatives in our task.

Although we mainly focus on face alignment of arbitrary views, our framework also naturally deals with partial occlusion and capable of inferring the visibility of landmarks. 

\section{Related Work}
\label{related}

Many approaches have been proposed for solving the face alignment problem.
The classic methods include the Active Appearance Model (AAM)~\cite{cootes2001active} and its variants.
AAM models faces by both appearance and shape information and optimises them in a holistic manner.
Facial parts detection based method~\cite{belhumeur2011localizing}, also known as Constrained Local Model, aim to detect facial parts with several part detectors, and use global constraints to estimate reasonable shapes.
Discriminative regression methods~\cite{cao2014face, xiong2013supervised, ren2014face, kazemi2014one} regress expected shape update based on local appearance information extracted from all current estimated landmarks.
Such methods have achieved state-of-the-art accuracy.
Our framework allows these state-of-the-art approaches to be applicable to non-frontal cases via recommendation over multiple models.
A number of existing studies propose to solve the multi-view face alignment problem \cite{zhou2005bayesian,xing2014towards}. These studies differ significantly from our work since they confine the yaw angle of faces within the range of $\sim\pm$45$\degree $ \cite{zhou2005bayesian}.
In~\cite{xing2014towards}, the dataset is chosen from the 300-W dataset~\cite{sagonas2013300} with an approximate range of $\sim\pm$45$\degree $, which is far smaller than the head pose range considered in this study.
%

Our work is closer to a few alignment studies that address the face alignment problem in a multimodal manner:

\noindent\textbf{Conditional Regression Forest \cite{dantone2012real}}. In this work, the facial points are estimated after a rough estimation of head pose.
The head pose is estimated according to the global appearance of the face image and regarded as the prior probability of each alignment model.
This method works quite well and gains improvement on traditional face alignment benchmarks, \eg~LFPW~\cite{belhumeur2011localizing} or LFW~\cite{dantone2012real}.
When this method is extended to dataset with extremely large pose variations, \eg~AFLW, the method fails due to erroneous head pose estimation, no matter how well each face alignment model is trained.
%

\noindent\textbf{Iterative Multi-Output Random Forests \cite{zhao2014unified}}. This study extends~\cite{dantone2012real} by iteratively re-estimating the head pose and facial expression based on the current estimated shape, and then re-estimating the facial landmarks with the updated prior model probability.
Further improvement is achieved in this work compared to~\cite{dantone2012real}.
Similar to~\cite{dantone2012real}, the algorithm begins with head pose estimation based on global appearance.
When applied to faces with large pose variation, if the initial head pose estimation is incorrect, the error is rather hard to be rectified during iterative optimisation since the initial shape is estimated with the wrong prior probability.

\noindent\textbf{Consensus of Regressors \cite{yu2014consensus}}. The study applies a mixture of models to handle different types of occluded faces.
The method works well on occlusion benchmarks like COFW~\cite{burgos2013robust}.
However, the method yields poor performance on profile-view cases because all regression models assume a limited pose variation.
From our experiments, we observe that even if models are provided to handle large pose variations, suboptimal results may be still generated due to the failure in drawing consensus amongst recommended landmarks with large spatial variations.
%

%
%

\noindent\textbf{Model Recommendation}. Model recommendation has been proposed for action recognition~\cite{matikainen2012model}. The study shows that a recommendation system is important for choosing one model out of a large set of possibilities. Their method follows the conventional collaborative filtering scheme, where user's rating is required on a small subset of models, and the goal is to use that subset of ratings to predict the ratings of remaining models. This is in contrast to our task where user rating is not available.

\section{Arbitrary-View Face Alignment}
\label{method}

\begin{figure}
  \centering
  \includegraphics[width=\linewidth]{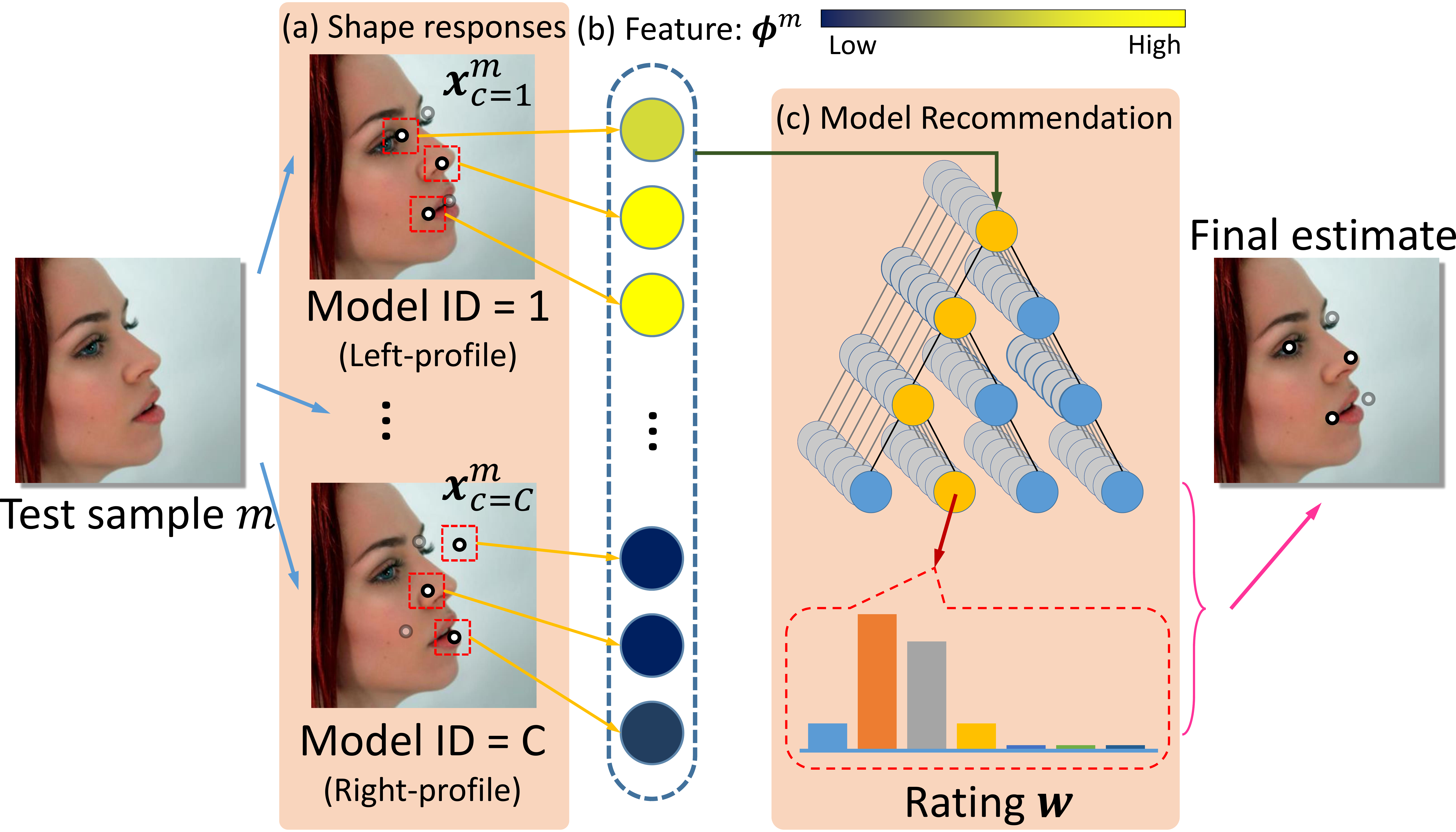}\\
  \caption{ There are three steps to align a face image with arbitrary pose: (a) extracting shape responses from \textit{model pool} (note that the model pool could cover the modeling of various in-plane or out-plane face rotation and external-object occlusion), (b) forming \textit{recommendation feature} vector based on model shape responses, and (c) model recommendation using the proposed \textit{Ensemble of Model Recommendation Trees}. }
  \label{fig:pipeline}
\end{figure}


Our aim is to predict the $N$ facial landmark locations given a face image with arbitrary view up to profile view, with potentially in-plane rotation, self- or external-object occlusion.
%
%
In addition to landmark locations, we also want to estimate a confidence value that indicates the visibility of each landmark.
%
%
%
%
%
Figure~\ref{fig:pipeline} provides an overview of using the proposed framework for aligning a face with arbitrary pose. We explain the training process involved in each step as follows. Table~\ref{tab:notation} summarises the notations we used in this study.

\begin{table}[t]
  \centering
  \begin{tabu}{l|l|l}
\tabucline[1.5pt]{-}
  Category               & Notation  & Meaning \\\hline
\multirow{5}{*}{Index} & $c = 1,...,C$ & Index of model         \\\cline{2-3}
                       & $m = 1,...,M$         & Index of sample         \\\cline{2-3}
                       & $n = 1,...,N$         & Index of facial landmark        \\\cline{2-3}
                       & $f = 1,...,F$         & Index of feature        \\\cline{2-3}
                       & $j$         &  Index of node in a tree       \\\hline
\multirow{5}{*}{\begin{tabular}[c]{@{}l@{}}Training\\ Set\end{tabular}}   & $\boldsymbol{x}^m_{c,n} \in \mathcal{X}, \mathbb{R}^2$        &Model shape responses       \\\cline{2-3}
                       & $\boldsymbol{y}^m_n \in \mathcal{Y}, \mathbb{R}^2$        & Shape ground truth       \\\cline{2-3}
                       & $\{m,n\} \in \mathcal{V}$         & Visibility ground truth       \\\cline{2-3}
                       & $\phi^m_f \in \mathcal{F}$         & Feature       \\\cline{2-3}
                       & $l^m_n \in \mathcal{L}$        & Label for classification \\\hline
\multirow{1}{*}{\begin{tabular}[c]{@{}l@{}}Model\end{tabular}} & $\boldsymbol{b}_c \in \{0,1\}^N$ & Model visibility protocol \\\hline
\multirow{1}{*}{\begin{tabular}[c]{@{}l@{}}Rating\end{tabular}} & $\boldsymbol{w}_j \in \mathbb{R}^{C}$ & Rating
\\\tabucline[1.5pt]{-}
\end{tabu}
    \caption{Mathematical notations used in this study.}
\label{tab:notation}
\end{table}


\noindent\textbf{a) Model pool construction}. We construct a \textit{model pool} of discriminative regression alignment models as depicted in Fig.~\ref{fig:pipeline}(a).
We assume there are $C$ models in the pool and $M$ training samples.
Each \textbf{model} $c$ is trained to accept a facial image and output the predicted landmarks
$\{\boldsymbol{x}_{c,n}\}^{N}_{n=1}$, where $c \in \left\{1,\dots,C\right\}$, and $N$ is the number of landmarks.
%
Here $\boldsymbol{x}$ is the $(x,y)$ coordinate of one landmark.
The landmarks predicted by all models over all samples collectively form the model shapes responses $\mathcal{X} = \{\boldsymbol{x}^m_{c,n}\}^{C,~~~N,~~~M}_{c=1,n=1,m=1}$. 

Each alignment model is trained to handle only a specific cluster of faces with specific view range and landmark visibility protocol\footnote{Variations in landmark visibility are caused by both self-occlusion due to extreme view pose, and external-object occlusion.}.
Models are different from each other, \eg~there can be a model specific to left-profile, or a model specific to 45$\degree$ out-of-plane rotated with mouth occluded.
We denote landmark protocol of model $c$ by a binary mask vector $\boldsymbol{b}_c \in \{0,1\}^{N}$, 1 for visible and 0 for invisible~(derived from ground truth visibility $\mathcal{V}$).
To train each model, we provide it with the labelled facial landmark locations $\mathcal{Y} = \{\boldsymbol{y}^m_n\}_{\{m,n\} \in \mathcal{V}}$ and visibility $\mathcal{V}$. Here $\{m,n\} \in \mathcal{V}$ only when the $n$-th landmarks in the $m$-th face is visible.
In other words, the training samples for a model must contain all visible landmarks defined by the associated model landmark protocol.
Note that we still force all the models to output the full set of landmarks, including those defined as invisible by the model, so as to facilitate the following recommendation process, \ie~$\boldsymbol{x}^m_{c,n} \in \mathcal{X}$ even if $(\boldsymbol{b}_c)_n = 0$.

Consequently, each alignment model is capable of handling its specific type of faces easily; the hard part lies in the way to infer the optimal model configuration given a face with potentially arbitrary pose and landmarks visibility. 

\noindent\textbf{b) Recommendation features extraction}.
All alignment models in the pool collectively produce the model shape responses $\mathcal{X}$ given a face image, as depicted in Fig.~\ref{fig:pipeline}(a).
From the given model shape responses we extract recommendation features as input to the EMRT.
Formally, we represent the recommendation feature of the $m$-th face image as $\boldsymbol{\phi}^m = \trans{(\phi_1^m,\dots,\phi_F^m)}$, where each dimension a facial part detection score of a visible landmark, and the dimension is denoted by $F = \sum_{c=1}^C \|\boldsymbol{b}_c\|_1$.
%
%
To obtain the part detection score, we apply a facial part detector (SIFT+SVM) on all visible landmarks defined in all models in the pool, and use the resulting detection scores to form the recommendation feature of a face.
Such feature is more discriminative than global appearance features used in the prior-head-pose-estimation method.
We collectively denote all features of $M$ training images as $\mathcal{F} = \{\phi^m_f\}^{M,~~~F}_{m=1,f=1}$.


\noindent\textbf{c) Model Recommendation by ensemble of recommendation trees}.
The core purpose of a recommendation system is to estimate a rating that describe the fitness of a model on a given task.
We denote a model rating vector as $\boldsymbol{w} \in \mathbb{R}^C$.
In our problem, the rating vector is used as a weight vector to combine the shape responses of different models in the pool to estimate the final landmark positions.
Similarly, the rating vector is also used to obtain the final visibility estimation.
The training details of the ensemble of recommendation trees are provided in Sec.~\ref{EMRT}.
Before we detail the training steps, we first review the conventional classification forest and discuss its limitations as a recommendation framework.


\subsection{Classification Forest for Model Recommendation}
\label{review}

A classification forest can be adapted for model recommendation by training it to pick up the most suitable model given the model shape responses obtained from the model pool. The rating may be approximated by using the posterior classification probability. Alternatively, a hard model selection can be done by finding the maximum votes from all trees.
In this section, we first give a brief review of classification forest and discuss its limitation as a model recommendation method.


Learning a conventional decision tree, \eg~the classification tree, typically starts from splitting the training set with optimal binary split function $h_\Theta(\boldsymbol{\phi}) \in \{0,1\}$, and repeats such process recursively until some criteria are reached.
Typically the form of $h$ is pre-defined, and for conventional classification tree, it could be represented by
\begin{equation}
\begin{aligned}
h_\Theta(\boldsymbol{\phi}) = &
\left\{
     \begin{array}{lr}
       0 & \phi_k \leq \tau \\
       1 & \phi_k > \tau
     \end{array}
   \right. .
\end{aligned}
\end{equation}
A sample is channeled to the left child if $h_\Theta(\boldsymbol{\phi}) = 0$, and to the right if otherwise.

To choose the best split function parameters $\Theta_j = \{k_j,\tau_j\}$ for node $j$, a set of candidate features and thresholds are randomly selected and evaluated. The optimal one that maximises the information gain is chosen as the node parameters.
%
%
%
Concretely, let denote the set of class labels as $\mathcal{L} = \{l^m\}^M_{m=1}$, where $l \in \{1,\dots,C\}$ is the model label. Given the current split node $j$ and its related sub training set $\mathcal{S}_j = \mathcal{F}_j \times \mathcal{L}_j \subset \mathcal{F} \times \mathcal{L}$, and let denote the two child nodes’ sub
training set by $\{\mathcal{S}_j^{(L)}, \mathcal{S}_j^{(R)}\}$, the information gain is defined as
\begin{equation}
I_j(\Theta_j) = H(\mathcal{S}_j) - \sum_{\theta \in \{L,R\}} \frac{|\mathcal{S}_j^{(\theta)}|}{|\mathcal{S}_j|} H(\mathcal{S}_j^{(\theta)})
\label{eqn:IG}
\end{equation}
In a classification setting, $H(\mathcal{S}_j)$ denotes the cross entropy of sub training set $\mathcal{S}_j$, represented as
\begin{equation}
H(\mathcal{S}_j) = \sum_{c = 1}^C -(p_j)_c \log (p_j)_c,
\label{eqn:entropy}
\end{equation}
where $p_j$ is the probability distribution calculated on $\mathcal{L}_j$.

Inference of classification forest can be achieved by feeding the feature $\boldsymbol{\phi}$ to all the trees and pass it downward according to the binary split function in each split node.
Each tree would output a predicted label proposed by the resulted leaf node.
We simply aggregate the decisions from all trees using a majority voting.

\begin{figure}
  \centering
  \includegraphics[width=0.95\linewidth]{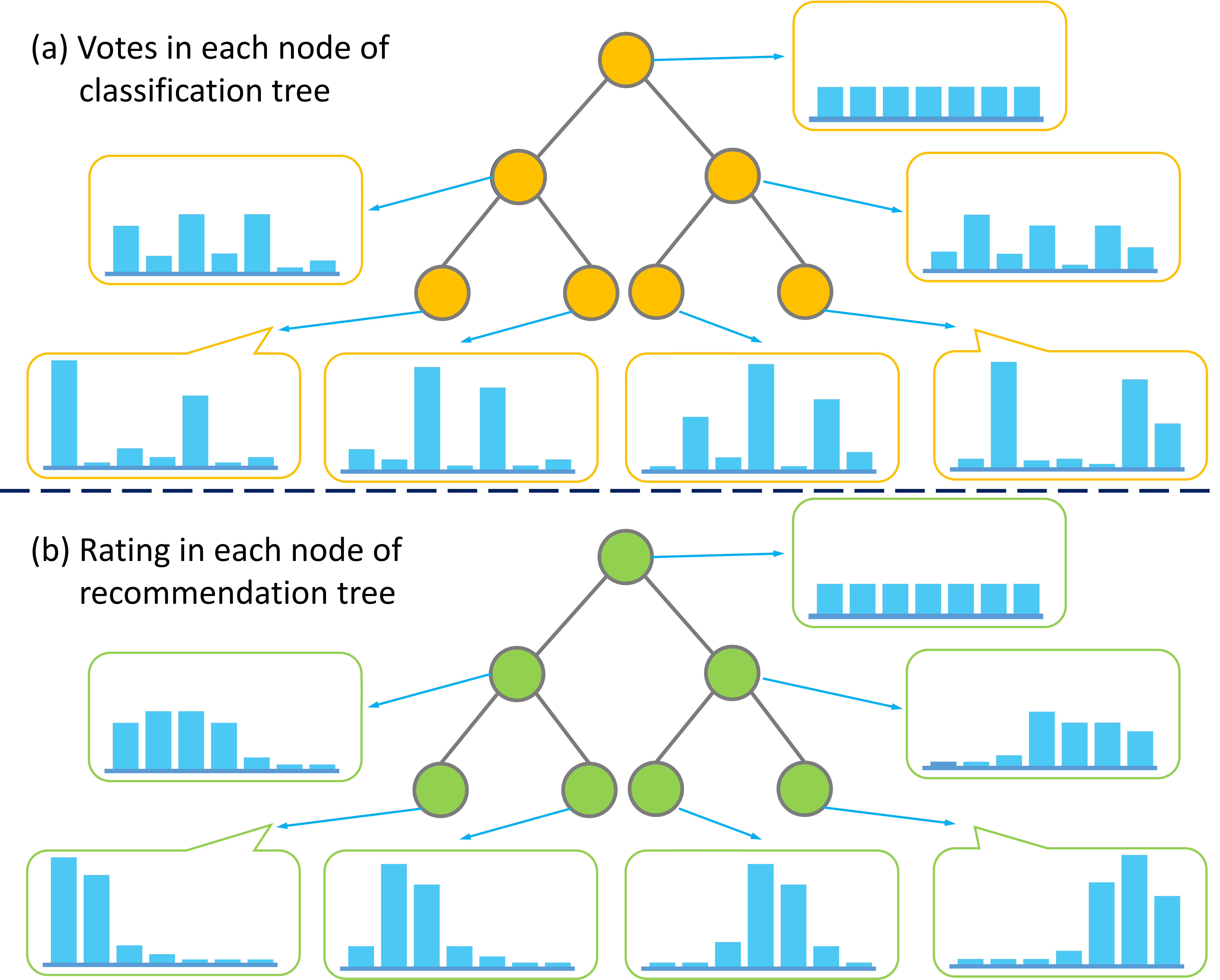}\\
  \caption{An illustration that highlights the differences between a conventional classification tree and a recommendation tree.
  To simplify the discussion, we assume the models in a pool are only yaw-specific without considering in-plane rotation or external-object occlusion. In each rating vector, the indices of rating are sorted based on the models' yaw angle.
  As can be observed, a classification tree neglects the potential relations between models, and may thus propose erroneous rating that advocates both left-profile-view and right-profile-view models.
  In contrast, the learning of a recommendation tree is directly supervised by facial landmark positions. Hence samples with similar facial shapes are naturally clustered, and hence similar models tend to share similar ratings. As a result, more robust and reasonable rating can be obtained.}
  \label{fig:split}
\end{figure}

Despite that the classification forest is effective in handling high-dimensional features of shape responses, it is not well-suited as a model recommendation framework for our problem, for the following reasons:

\noindent
(1) The rating generated using the posterior class probability of a classification forest tends to simultaneously recommend non-related models far away from each other in the model space (\eg~left-profile and right-profile models), as depicted in Fig.~\ref{fig:split}(a). Such a poor recommendation will lead to erroneous landmark estimation.
The poor recommendation is mainly caused by the splitting function of classification forest, in which the cross entropy formulation (Eq.~(\ref{eqn:entropy})) does not naturally distinguish the different models.
This drawback is resolved with our new splitting function. Fig.~\ref{fig:split} shows an illustration that highlights the difference between the ratings generated by the classification tree and a recommendation tree. The rating vector generated by the recommendation tree tends to recommend models close in the shape space.
We show in Sec.~\ref{exp} that the proposed recommendation trees outperform the classification trees.

\noindent
(2) The classification forest does not fully exploit the available supervision signal, namely landmark positions $\mathcal{Y}$ and visibility $\mathcal{V}$ in the training process.


\subsection{Learning Ensemble of Model Recommendation Trees}
\label{EMRT}

We wish to address the shortcomings of classification forest through clustering similar faces naturally with a specially designed split mechanism.
There are three main differences between the learning of a recommendation tree and a classification tree: (1) In a recommendation tree, the splitting of samples are directly related to their facial landmark locations instead of the expected model index as classification target.
(2) To learn a recommendation tree, we need to provide the full supervision signal $\{\mathcal{F}, \mathcal{X}, \mathcal{Y}, \mathcal{V}\}$ rather than just $\mathcal{F} \times \mathcal{L}$, as in the classification forest.
(3)
A recommendation tree incorporates the rating vector $\boldsymbol{w}_j \in \mathbb{R}^C$ into the optimisation of each node $j$.
%
%
All these differences are introduced aiming to cluster training samples with similar rating to the same tree node, and the similarity of rating is computed purely based on $\mathcal{X}$ and $\mathcal{Y}$.
%


More precisely, given current split node $j$ and its related sub training set $\mathcal{S}_j = \{\mathcal{F}_j, \mathcal{X}_j, \mathcal{Y}_j, \mathcal{V}\}$,
we define a new training objective as follows
\begin{equation}
I_j(\Theta_j,\boldsymbol{w}_j^{(\theta)}) = H(\mathcal{S}_j,\boldsymbol{w}_j) - \sum_{\theta \in \{L,R\}} \frac{|\mathcal{S}_j^{(\theta)}|}{|\mathcal{S}_j|} H(\mathcal{S}_j^{(\theta)},\boldsymbol{w}_j^{(\theta)}) ,
\label{eqn:IGnew}
\end{equation}
where $H(\mathcal{S}_j,\boldsymbol{w}_j)$ is defined as
\begin{equation}
H(\mathcal{S}_j,\boldsymbol{w}_j) = \sum_{\substack{ m,n: \\ \boldsymbol{x}^m \in \mathcal{S}_j, \\ \{m,n\} \in \mathcal{V} }}
\| \boldsymbol{y}^m_n - \sum_{c=1}^C w_{jc} \boldsymbol{x}^m_{c,n} \|_2^2.
\end{equation}
Note that both $\mathcal{Y}$ and $\mathcal{V}$ are directly involved in the optimisation function, in contrast to the conventional classification forest.
Importantly, the information gain is maximised not only based on $\Theta_j$, but also the rating vectors in the parent and child nodes $\{\boldsymbol{w}_j,\boldsymbol{w}_j^{(L)},\boldsymbol{w}_j^{(R)}\}$.
%

Optimisation with Eq.~(\ref{eqn:IGnew}) is more complex than Eq.~\ref{eqn:IG}, which is used in the conventional classification forest.
In the classification forest, optimisation is only performed over the split parameters $\Theta_j$, where we just enumerate several candidate features and thresholds to find the optimal $\Theta_j$.
On the contrary, the optimisation in the recommendation forest needs to account for both the split parameters $\Theta_j$ and rating $\boldsymbol{w}_j^{(\theta)}(\Theta_j), \theta \in \{L,R\}$.
In particular, for each candidate feature-threshold pair, we need to find the optimal $\boldsymbol{w}_j^{(\theta)}(\Theta_j), \theta \in \{L,R\}$.
%
%
Assume that the $\Theta_j$ is given, the partition of $\mathcal{S}_j$ is determined, the optimal $\boldsymbol{w}_j^{(\theta)}(\Theta_j)$ can be obtained by
\begin{equation}
\begin{aligned}
\boldsymbol{w}_j^{(\theta)}(\Theta_j) = \argmin_{\boldsymbol{w}} & \sum_{\substack{ m,n: \\ \boldsymbol{x}^m \in \mathcal{S}_j^{(\theta)}, \\ \{m,n\} \in \mathcal{V} }}
\| \boldsymbol{y}^m_n - \sum_{c=1}^C w_{c} \boldsymbol{x}^m_{c,n} \|_2^2 \\
s.t. ~~~~ \boldsymbol{w} & \geq 0,
\|\boldsymbol{w}\|_1 = 1,
\end{aligned}
\label{eqn:inneropt}
\end{equation}
which can be efficiently solved by Quadratic Programming packages.
Note that the regularity constraints over $\boldsymbol{w}$ in Eq. \ref{eqn:inneropt} is important to avoid unreasonable recommendation results and to enforce facial shape constraints.



\subsection{Landmark Inference with EMRT}
\label{inference}

Despite the learning procedure of the proposed recommendation tree is more sophisticated than a traditional tree, the inference efficiency is identical to the conventional tree.
Specifically, given an image in the testing stage, we first extract the shape responses $\tilde{\boldsymbol{x}}_{c,n}$ and subsequently compute the recommendation feature $\tilde{\boldsymbol{\phi}}$. The feature is fed to a recommendation tree and passed down from the root node to the leaf node according to the binary split function $h_{\Theta_j}(\tilde{\boldsymbol{\phi}})$ stored in each node.
For each tree, rating $\boldsymbol{w}$ stored in the reached leaf node is used to estimate the predicted landmark locations $\tilde{\boldsymbol{y}}_n$ and visibility confidence $\tilde{v}_n$,
\begin{equation}
\begin{aligned}
\tilde{\boldsymbol{y}}_n = \sum_{c=1}^C w_c \tilde{\boldsymbol{x}}_{c,n},~~~~~~\tilde{v}_n = \sum_{c: (\boldsymbol{b}_c)_n = 1} w_c \tilde{\phi}_{c,n},
\end{aligned}
\label{eqn:inference}
\end{equation}
where $\tilde{\phi}_{c,n}$ is retrieved from the $F$-dimensional $\tilde{\boldsymbol{\phi}}$.

To obtain the final estimation from different trees, we average all the predictions (landmark locations and visibility confidence) weighted by the number of training samples in the reached leaf node of each tree. Visibility of a landmark is obtained via comparing the estimated confidence with a threshold $\gamma$.

\section{Experiments}
\label{exp}

\paragraph{Datasets.}
Following the objective of this study of aligning arbitrary-view face up to profile view, we choose the following datasets with extremely large pose variations. 

\textit{AFLW dataset} \cite{kostinger2011annotated}: This dataset was collected from Flickr hence exhibits in-the-wild setting. The original dataset contains 25,993 images, of which 27.48\% have yaw angle larger than 45$\degree$ and 3.83\% larger than 90$\degree$. The large angle range and in-the-wild setting is challenging for common methods without specifically designed mechanism.
Each face is annotated with at most 21 landmarks. 
Similar to~\cite{yang2013sieving, belhumeur2011localizing}, we discard two ear landmarks to enforce more stable face shape. Non-visible landmark annotation, including both self-occlusion and external-object-occlusion, is not provided.
To our knowledge, our study is the first attempt to train and evaluate on this challenging dataset without removing non-frontal faces.

%
%
%
%
%
%

\textit{MultiPIE dataset} \cite{sim2003cmu}: This dataset was originally collected in a constrained lab environment with a setting of multiple poses, illuminance and facial expressions.
A subset of 6152 faces are defined~\cite{sim2003cmu} and labelled with either 39 (profile-view) or 68 (frontal-view) facial landmarks.
Zhu~\etal~\cite{zhu2012face} further annotate another 400 profile-view faces with 39 landmarks.
In our experiments, these two subsets of data are merged and cleaned to form a 6547-image dataset.

\textit{AFW dataset} \cite{zhu2012face}: This original dataset contains 468 faces with in-the-wild setting.
Each face is annotated with at most 6 landmarks. Non-visible landmark annotation is not provided similar to AFLW.
We note that a subset of AFW, totally 337 faces, were selected and re-annotated with 68 landmarks by~\cite{sagonas2013300}.
These 337 faces are frontally biased, hence we still employ the original 468 faces and the associated annotations.

\noindent\textbf{Evaluation metric.}
We apply percentage of mean squared error normalised by the upright distance between the highest eyebrow landmark and the chin landmark as the evaluation metric.
Errors are calculated on all annotated landmarks, excluding non-annotated or invisible landmarks, since no ground truths are available.
We do not normalise error by inter-ocular distance because one of the two eyes is invisible for many profile view cases.
%


\subsection{Comparison with Existing Methods}
\label{exp:soa}

\paragraph{Comparison with \cite{zhu2012face}.} To our knowledge, \cite{zhu2012face} is one of the most recent works that is capable of handling profile-view and frontal-view faces simultaneously.
Model of \cite{zhu2012face} is trained on only 900 images of MultiPIE dataset, which is a rather small fraction of the original dataset.
Since the training of a discriminative regression model requires more training samples, we employ the full MultiPIE dataset and perform evaluations by averaging performance on 5-fold cross validation.
For a fair comparison, we re-train the model of \cite{zhu2012face} in the same way using their released codes.
Note that each person might contribute multiple facial images. We ensure that the images from the same person only appear within the same fold.

Similar to \cite{zhu2012face}, we further evaluate the models trained on the full MultiPIE set on the \textbf{full} AFW dataset.
As in \cite{zhu2012face}, we use a trained linear mapping to obtain the 6-point estimate by each method.
Since \cite{zhu2012face} predefines the mixture of trees, it is infeasible to train the model of \cite{zhu2012face} with AFLW landmarks protocol. Hence we do not compare with \cite{zhu2012face} on the AFLW dataset.

\begin{table}[h]
  \centering
  \begin{tabu}{lcc}
\tabucline[1.5pt]{-}
Approach                    & MultiPIE & AFW \\\hline
\cite{zhu2012face}                     & 2.74                                              & 4.68                                                 \\\hline
Ours      & \textbf{2.12}                                     & \textbf{3.55}\\\tabucline[1.5pt]{-}
\end{tabu}
  \caption{Mean error comparison with \cite{zhu2012face} on two datasets (also applied in \cite{zhu2012face}).}\label{tab:multipie}
\end{table}

\begin{figure}[t]
\centering
\subfigure[Results from MultiPIE.]{\label{fig:a}\includegraphics[width=0.49\linewidth]{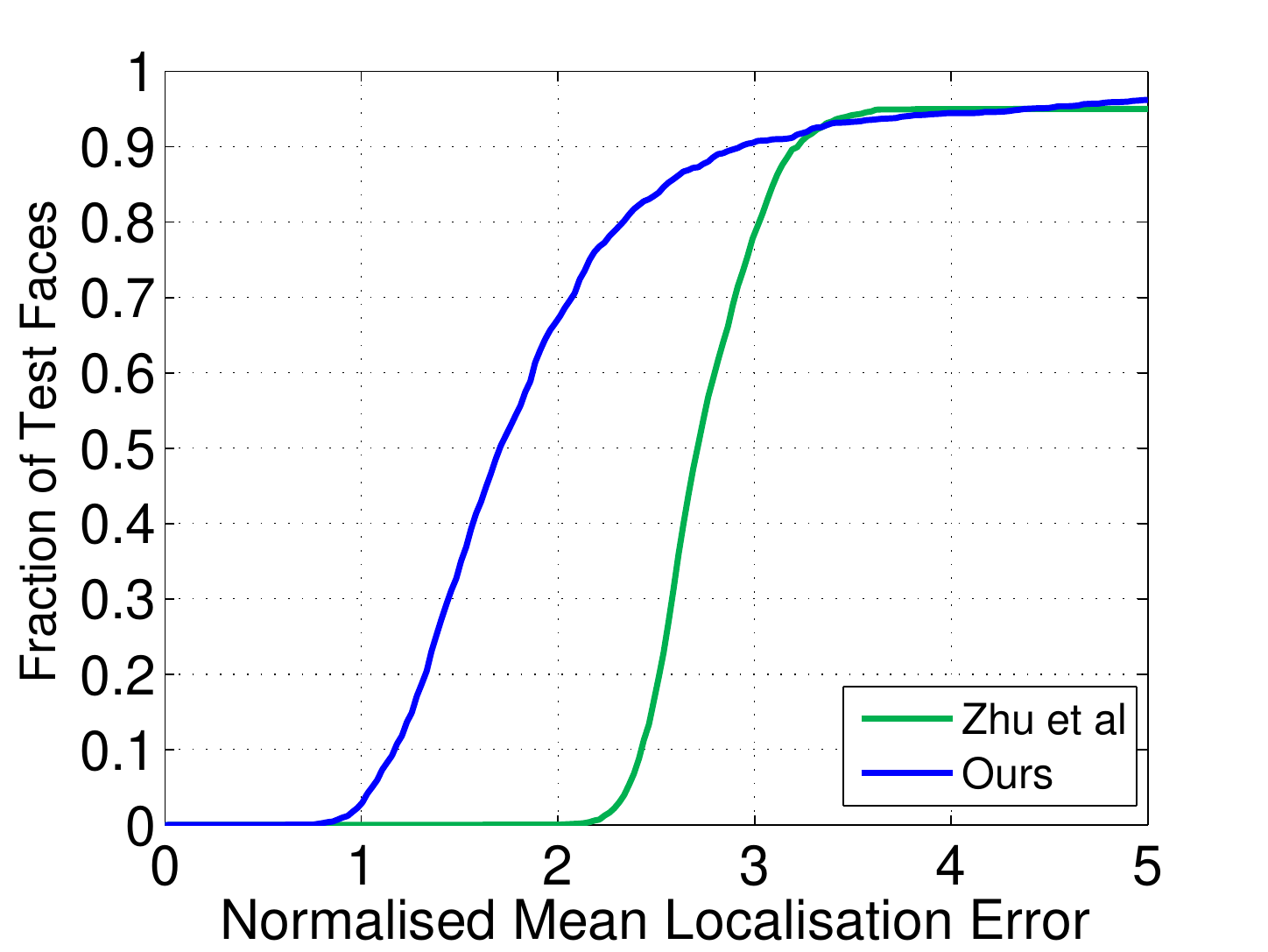}}
\subfigure[Results from AFW.]{\label{fig:b}\includegraphics[width=0.49\linewidth]{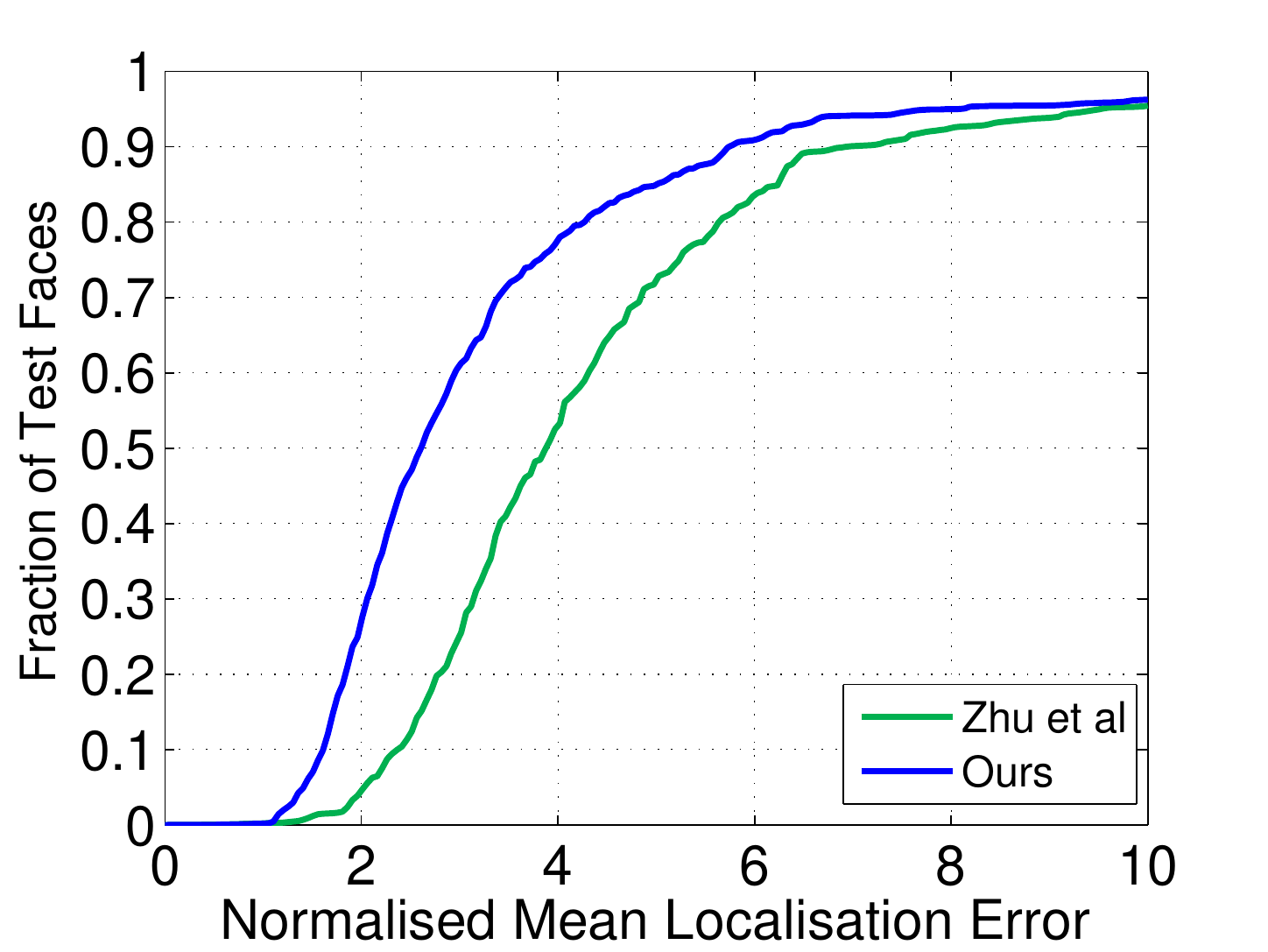}}
\caption{CED curve compared with \cite{zhu2012face} on MultiPIE and AFW.}
\label{fig:ced}
\end{figure}

\begin{figure*}
  \centering
  \includegraphics[width=0.95\linewidth]{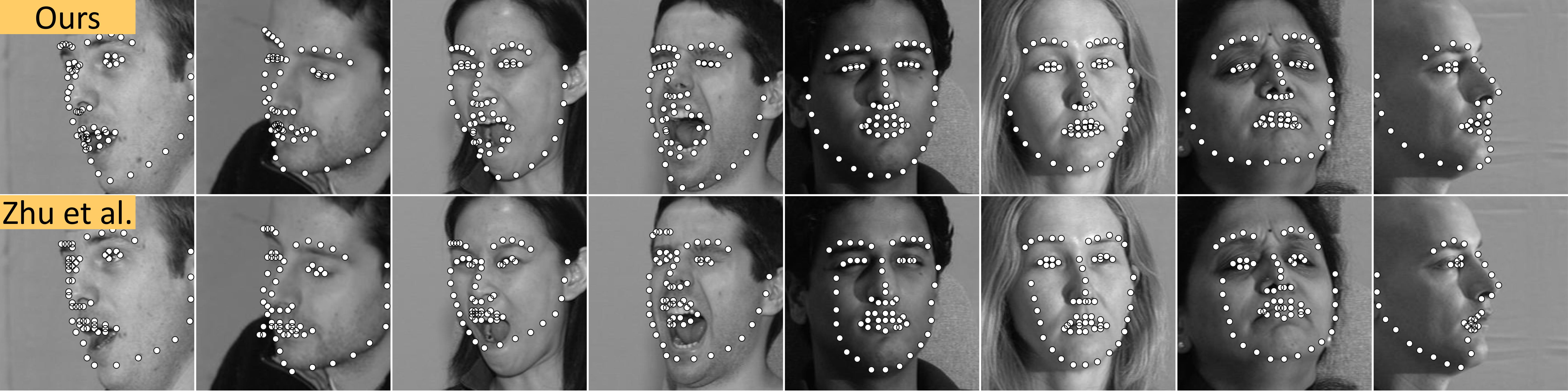}\\
  \caption{Comparison of representative results between our method (first row) and \cite{zhu2012face} (second row).}
  \label{fig:multipieshowoff}
\end{figure*}

Results of the comparison are reported in Table~\ref{tab:multipie}.
We also compare the relative cumulative error distribution (CED) curve for the two methods on MultiPIE and AFW datasets in Fig.~\ref{fig:ced}.
Based on the results, it is observed that our framework outperforms \cite{zhu2012face} on both the datasets applied in their work. More significant improvement is observed on AFW dataset with in-the-wild setting.  The results suggest the effectiveness of using  the discriminative method along with the proposed recommendation framework for handling unconstrained settings.
We provide qualitative examples in Fig~.\ref{fig:multipieshowoff} comparing against~\cite{zhu2012face}.

\noindent\textbf{Comparison with \cite{xiong2013supervised, ren2014face, kazemi2014one}.} The methods proposed in these three studies are representative discriminative regression approaches. They have achieved state-of-the-art results on commonly used frontal-view-biased datasets.
In this part we compare our proposed recommendation framework with these regression methods to show the benefits of adopting a model recommendation framework.
Since the training codes of~\cite{xiong2013supervised, ren2014face} are not publicly available, we evaluate all these methods based on our re-implementation.
Our re-implementation achieves comparable performance on common benchmarks (\eg~LFPW~\cite{belhumeur2011localizing}, Helen~\cite{le2012interactive}) with theirs reported in the literature.

We evaluate our approach and the three baseline methods on the \textbf{full} AFLW dataset since it is more challenging than the MultiPIE dataset.
Since AFLW also does not provide a train-test set partition, we perform a 5-fold cross-validation and report the averaged performance across all the folds.
%

\begin{table}[h]
  \centering
  \begin{tabu}{lc}
\tabucline[1.5pt]{-}
Approach                    & Mean error over AFLW dataset \\\hline
SDM~\cite{xiong2013supervised}                     & 6.98                                                                                             \\\hline
LBF~\cite{ren2014face}                     & 7.47                                                                                            \\\hline
ERT~\cite{kazemi2014one}                     & 7.70                                                                                             \\\hline
Ours      & \textbf{4.01}                                    \\\tabucline[1.5pt]{-}
\end{tabu}
  \caption{Mean error comparison with \cite{xiong2013supervised, ren2014face, kazemi2014one} on the full AFLW dataset.}\label{tab:aflw}
\end{table}

Note that we only train the discriminative regression methods with the frontal face subset.
Training the baseline methods using the full training set is infeasible since AFLW contains varying point sets and these methods cannot handle this issue. Simply filling the missing points with transformed mean shape is possible, but training the baseline methods with this augmented point sets lead to worse performance than using only the frontal face subset.
%
%

\begin{figure*}
  \centering
  \includegraphics[width=0.95\linewidth]{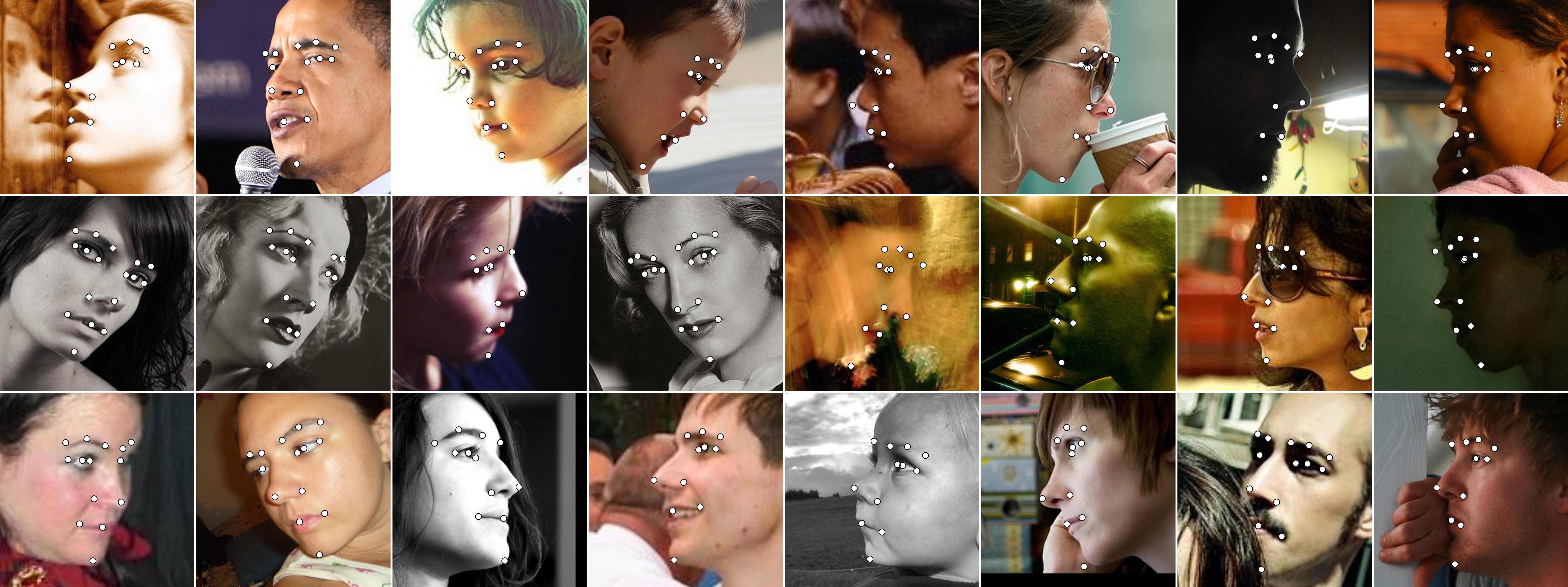}\\
  \caption{Illustrative results by our method on AFLW dataset. It is worth pointing that all faces have extreme view pose (up to profile view or with in-plane rotation).}
  \label{fig:aflwshowoff}
\end{figure*}

The comparative results are summarised in Table~\ref{tab:aflw}.
The results suggest the importance of deploying multiple models together with a model recommendation framework to handle arbitrary views.
%
Qualitative results by our method are shown in Fig.~\ref{fig:aflwshowoff}.

\subsection{Further Analysis}
\label{exp:val}

To validate the effectiveness of the proposed model recommendation trees, we further compare our approach with some variants derived from our framework and alternative strategies:
\begin{enumerate}[(a)]
  \item No recommendation at all, \ie~always choosing the frontal model as recommended model. This variant is equivalent to SDM~\cite{xiong2013supervised};
  \item Prior head pose estimation before face alignment. This method is conceptually similar to the conditional regression forest~\cite{dantone2012real};
  \item Optimising landmarks locations based on multi-response map. This method is conceptually similar to the study on consensus by regressors~\cite{yu2014consensus};
  \item Classification forest choosing the model with top voting;
  \item Classification forest using posterior class probability as rating.
\end{enumerate}

Note that all the baselines exploits the same SDM models as those used in our approach to ensure a fair comparison in the model recommendation process.
We compare all the methods on the challenging full AFLW dataset and the training setting is similar to that described in Sec.~\ref{exp:soa}.
We adopt three evaluation metrics, namely mean normalised error, visibility detection accuracy, and the corresponding average precision.
Visibility detection accuracy refers to the percentage of landmarks predicted with correct visibility.

\begin{figure}
  \centering
  \includegraphics[width=\linewidth]{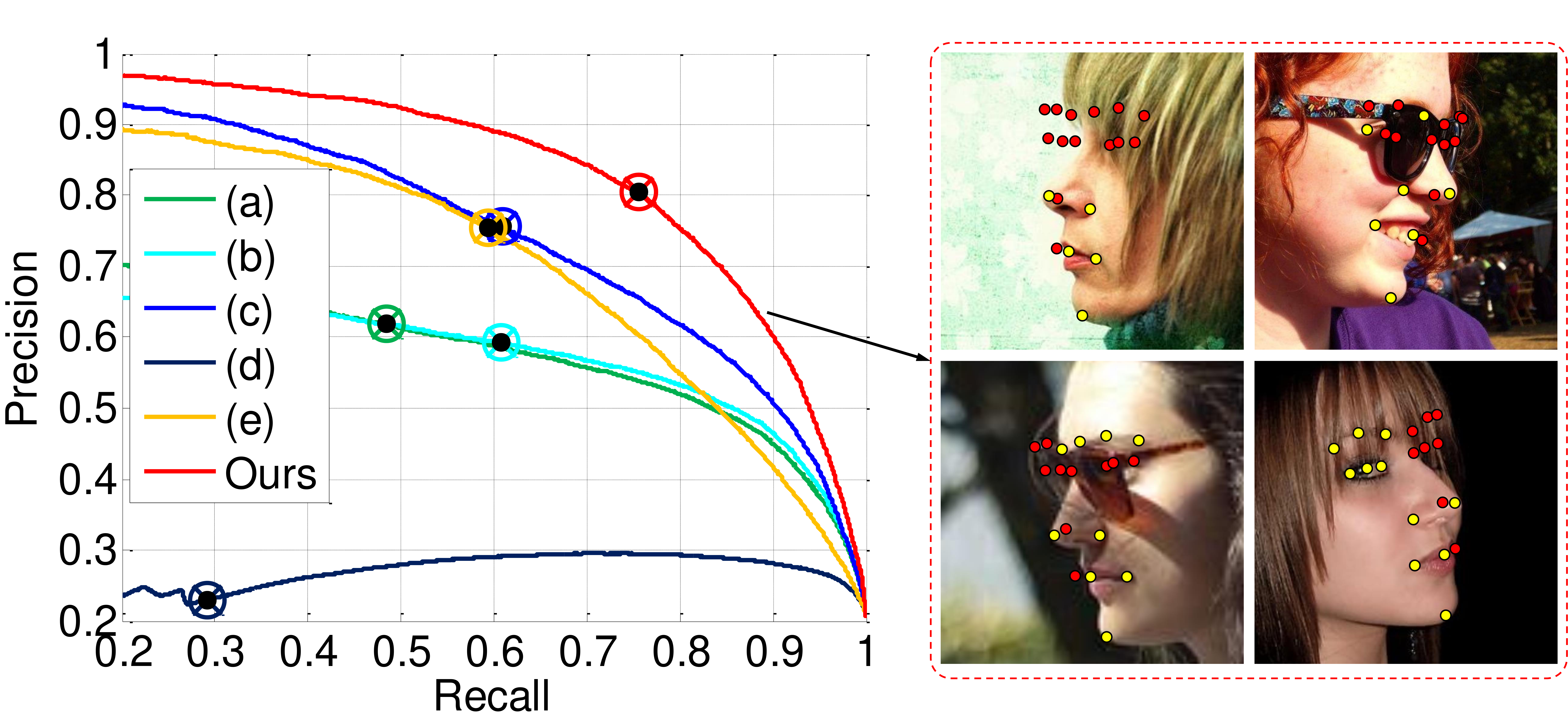}\\
  \caption{Visibility detection Precision-Recall curve compared to various baseline methods, with representative samples. The black dot on the curve denotes the visibility threshold. Threshold for each method is chosen to maximise visibility detection accuracy on the validation set. Colors of the annotated landmarks indicate our estimate of visibility (yellow for visible and red for invisible). Best viewed in color.}
  \label{fig:occl}
\end{figure}

Results are provided in Table~\ref{tab:val}.
Precision-Recall curve of visibility evaluation is depicted in Fig.~\ref{fig:occl}.
Following~\cite{burgos2013robust}, we treat invisible landmarks as `positive'.
The results suggest the effectiveness of the the proposed model recommendation trees.
It can be observed that an application of approaches with prior head pose estimation ~\cite{dantone2012real} or occlusion handling~\cite{yu2014consensus} only gains marginal improvement compared with the frontal-face SDM on the challenging arbitrary-view face alignment task.
Moreover, a naive application of classification forest (both choosing top voted model or using posterior class probability as rating) cannot achieve the desired results.
%

\begin{table}[h]
  \centering
  \begin{tabu}{cccc}                                                                                              \\\tabucline[1.5pt]{-}
                & \begin{tabular}[c]{@{}c@{}} Mean error \\on AFLW \end{tabular}  & \begin{tabular}[c]{@{}c@{}} Visibility detection \\ accuracy (\%) \end{tabular} & \begin{tabular}[c]{@{}c@{}} Visibility \\ AP (\%) \end{tabular} \\\hline
(a)                    & 6.98                                                        & 83.20  & 42.35                                   \\\hline
(b)                    & 6.61                                                                            & 83.33  & 41.78               \\\hline
(c)                     & 6.65                                                                                  & 87.89   & 47.08       \\\hline
(d)                     & 9.24                                                                       & 65.26  & 25.75                    \\\hline
(e)                     & 7.62                                                                     & 87.65  & 45.79                    \\\hline
Ours      & \textbf{4.01}                                   & \textbf{91.18} & \textbf{49.35} \\\tabucline[1.5pt]{-}
\end{tabu}
  \caption{Comparison of model recommendation between the proposed recommendation trees and various baseline methods.}\label{tab:val}
\end{table}

%
%

\section{Conclusion}

We have presented a novel model recommendation framework to simultaneously handle face alignment of arbitrary-view faces up to profile view. The framework can flexibly work with different existing discriminative regression based methods. It is capable of accommodating different landmark protocols with varying point sets and inferring visibility of landmarks.
Extensive experiments on the full set of AFLW, MultiPIE, and AFW demonstrate the effectiveness of our approach over existing multimodal face alignment methods.

{\small
\bibliographystyle{ieee}
\bibliography{short,rcm}

\begin{thebibliography}{10}\itemsep=-1pt

\bibitem{belhumeur2011localizing}
P.~N. Belhumeur, D.~W. Jacobs, D.~Kriegman, and N.~Kumar.
\newblock Localizing parts of faces using a consensus of exemplars.
\newblock In {\em CVPR}, pages 545--552, 2011.

\bibitem{breiman2001random}
L.~Breiman.
\newblock Random forests.
\newblock {\em Machine learning}, 45(1):5--32, 2001.

\bibitem{burgos2013robust}
X.~P. Burgos-Artizzu, P.~Perona, and P.~Doll{\'a}r.
\newblock Robust face landmark estimation under occlusion.
\newblock In {\em ICCV}, pages 1513--1520, 2013.

\bibitem{cao2014face}
X.~Cao, Y.~Wei, F.~Wen, and J.~Sun.
\newblock Face alignment by explicit shape regression.
\newblock {\em IJCV}, 107(2):177--190, 2014.

\bibitem{chen2013automatic}
C.~Chen, A.~Dantcheva, and A.~Ross.
\newblock Automatic facial makeup detection with application in face
  recognition.
\newblock In {\em ICB}, pages 1--8, 2013.

\bibitem{cootes2001active}
T.~F. Cootes, G.~J. Edwards, and C.~J. Taylor.
\newblock Active appearance models.
\newblock {\em TPAMI}, 23(6):681--685, 2001.

\bibitem{cootes2012robust}
T.~F. Cootes, M.~C. Ionita, C.~Lindner, and P.~Sauer.
\newblock Robust and accurate shape model fitting using random forest
  regression voting.
\newblock In {\em Computer Vision--ECCV 2012}, pages 278--291. Springer, 2012.

\bibitem{dantone2012real}
M.~Dantone, J.~Gall, G.~Fanelli, and L.~Van~Gool.
\newblock Real-time facial feature detection using conditional regression
  forests.
\newblock In {\em CVPR}, pages 2578--2585, 2012.

\bibitem{datta2011hierarchical}
A.~Datta, R.~Feris, and D.~Vaquero.
\newblock Hierarchical ranking of facial attributes.
\newblock In {\em AFGR}, pages 36--42. IEEE, 2011.

\bibitem{dou2014benchmarking}
P.~Dou, Y.~Wu, S.~K. Shah, and I.~A. Kakadiaris.
\newblock Benchmarking 3d pose estimation for face recognition.
\newblock In {\em Pattern Recognition (ICPR), 2014 22nd International
  Conference on}, pages 190--195. IEEE, 2014.

\bibitem{du2014compound}
S.~Du, Y.~Tao, and A.~M. Martinez.
\newblock Compound facial expressions of emotion.
\newblock {\em Proceedings of the National Academy of Sciences},
  111(15):E1454--E1462, 2014.

\bibitem{hermosilla2011thermal}
G.~Hermosilla, P.~Loncomilla, and J.~Ruiz-del Solar.
\newblock Thermal face recognition using local interest points and descriptors
  for hri applications.
\newblock In {\em RoboCup 2010: Robot Soccer World Cup XIV}, pages 25--35.
  Springer, 2011.

\bibitem{kazemi2014one}
V.~Kazemi and S.~Josephine.
\newblock One millisecond face alignment with an ensemble of regression trees.
\newblock In {\em CVPR}, 2014.

\bibitem{kostinger2011annotated}
M.~Kostinger, P.~Wohlhart, P.~M. Roth, and H.~Bischof.
\newblock Annotated facial landmarks in the wild: A large-scale, real-world
  database for facial landmark localization.
\newblock In {\em ICCVW}, pages 2144--2151, 2011.

\bibitem{le2012interactive}
V.~Le, J.~Brandt, Z.~Lin, L.~Bourdev, and T.~S. Huang.
\newblock Interactive facial feature localization.
\newblock In {\em ECCV}, pages 679--692, 2012.

\bibitem{matikainen2012model}
P.~Matikainen, R.~Sukthankar, and M.~Hebert.
\newblock Model recommendation for action recognition.
\newblock In {\em CVPR}, pages 2256--2263, 2012.

\bibitem{ren2014face}
S.~Ren, X.~Cao, Y.~Wei, and J.~Sun.
\newblock Face alignment at 3000 fps via regressing local binary features.
\newblock In {\em CVPR}, 2014.

\bibitem{sagonas2013300}
C.~Sagonas, G.~Tzimiropoulos, S.~Zafeiriou, and M.~Pantic.
\newblock 300 faces in-the-wild challenge: The first facial landmark
  localization challenge.
\newblock In {\em ICCVW}, pages 397--403, 2013.

\bibitem{shen2013detecting}
X.~Shen, Z.~Lin, J.~Brandt, and Y.~Wu.
\newblock Detecting and aligning faces by image retrieval.
\newblock In {\em Computer Vision and Pattern Recognition (CVPR), 2013 IEEE
  Conference on}, pages 3460--3467. IEEE, 2013.

\bibitem{sim2003cmu}
T.~Sim, S.~Baker, and M.~Bsat.
\newblock The cmu pose, illumination, and expression database.
\newblock {\em TPAMI}, 25(12):1615--1618, 2003.

\bibitem{sun2014deep}
Y.~Sun, Y.~Chen, X.~Wang, and X.~Tang.
\newblock Deep learning face representation by joint
  identification-verification.
\newblock In {\em Advances in Neural Information Processing Systems}, pages
  1988--1996, 2014.

\bibitem{taigman2014deepface}
Y.~Taigman, M.~Yang, M.~Ranzato, and L.~Wolf.
\newblock {DeepFace}: Closing the gap to human-level performance in face
  verification.
\newblock In {\em CVPR}, 2014.

\bibitem{xing2014towards}
J.~Xing, Z.~Niu, J.~Huang, W.~Hu, and S.~Yan.
\newblock Towards multi-view and partially-occluded face alignment.
\newblock In {\em CVPR}, pages 1829--1836, 2014.

\bibitem{xiong2013supervised}
X.~Xiong and F.~De~la Torre.
\newblock Supervised descent method and its applications to face alignment.
\newblock In {\em CVPR}, pages 532--539, 2013.

\bibitem{yang2013structured}
C.-Y. Yang, S.~Liu, and M.-H. Yang.
\newblock Structured face hallucination.
\newblock In {\em Computer Vision and Pattern Recognition (CVPR), 2013 IEEE
  Conference on}, pages 1099--1106. IEEE, 2013.

\bibitem{yang2013sieving}
H.~Yang and I.~Patras.
\newblock Sieving regression forest votes for facial feature detection in the
  wild.
\newblock In {\em Computer Vision (ICCV), 2013 IEEE International Conference
  on}, pages 1936--1943. IEEE, 2013.

\bibitem{yu2014consensus}
X.~Yu, Z.~Lin, J.~Brandt, and D.~N. Metaxas.
\newblock Consensus of regression for occlusion-robust facial feature
  localization.
\newblock In {\em ECCV}, pages 105--118, 2014.

\bibitem{zen2014unsupervised}
G.~Zen, E.~Sangineto, E.~Ricci, and N.~Sebe.
\newblock Unsupervised domain adaptation for personalized facial emotion
  recognition.
\newblock In {\em Proceedings of the 16th International Conference on
  Multimodal Interaction}, pages 128--135. ACM, 2014.

\bibitem{Zhang2014}
Z.~Zhang, P.~Luo, C.~C. Loy, and X.~Tang.
\newblock Facial landmark detection by deep multi-task learning.
\newblock In {\em ECCV}, pages 94--108, 2014.

\bibitem{zhao2014unified}
X.~Zhao, T.-K. Kim, and W.~Luo.
\newblock Unified face analysis by iterative multi-output random forests.
\newblock In {\em CVPR}, pages 1765--1772, 2014.

\bibitem{zhou2005bayesian}
Y.~Zhou, W.~Zhang, X.~Tang, and H.~Shum.
\newblock A {B}ayesian mixture model for multi-view face alignment.
\newblock In {\em CVPR}, volume~2, pages 741--746. IEEE, 2005.

\bibitem{zhu2012face}
X.~Zhu and D.~Ramanan.
\newblock Face detection, pose estimation, and landmark localization in the
  wild.
\newblock In {\em CVPR}, pages 2879--2886, 2012.

\end{thebibliography}
}

\end{document}